\PassOptionsToPackage{table}{xcolor}
\documentclass[webpdf,modern,large,namedate]{oup-authoring-template}

\usepackage{multirow}
\usepackage{xcolor}
\usepackage{booktabs}

\graphicspath{{Fig/}}
\newcommand{\societylogo}{}

\theoremstyle{thmstyleone}%
\theoremstyle{thmstyletwo}%
\theoremstyle{thmstylethree}%

\begin{document}

\journaltitle{Bioinformatics}
\DOI{DOI HERE}
\copyrightyear{2026}
\pubyear{2026}
\access{Advance Access Publication Date: Day Month Year}
\appnotes{Application Note}

\firstpage{1}


\title[COMPASS]{COMPASS: Comet Object Measurement Pipeline with Automated Selection and Scoring}

\author[1]{Jack Roberts}
\author[2]{Canya Lu}
\author[3]{Alexis Michelle Lawson}
\author[4]{Kaitlyn Holden}
\author[3]{Gerald S. Wilkinson}
\author[4,$\ast$]{Anne M. Bronikowski}
\author[1,5,$\ast$]{Ritambhara Singh}

\address[1]{\orgdiv{Department of Computer Science}, \orgname{Brown University}, \orgaddress{Providence, RI, USA}}

\address[2]{\orgdiv{Department of Computer Science}, \orgname{Bryn Mawr College}, \orgaddress{Bryn Mawr, PA, USA}}

\address[3]{\orgdiv{Department of Biology}, \orgname{University of Maryland}, \orgaddress{College Park, MD, USA}}

\address[4]{\orgdiv{Department of Integrative Biology, W. K. Kellogg Biological Station, Michigan State University, Hickory Corners, MI, USA}}

\address[5]{\orgdiv{Department of Computing and Data Science}, \orgname{University of Illinois Urbana-Champaign}, \orgaddress{Urbana, IL, USA}}


\corresp[$\ast$]{Co-corresponding authors.}

\received{Date}{0}{Year}
\revised{Date}{0}{Year}
\accepted{Date}{0}{Year}



\abstract{
\mdseries
\textbf{Summary:}
The single-cell gel electrophoresis ("comet") assay is a widely used technique for quantifying DNA damage at the individual cell level. However, image analysis often relies on manual inspection or semi-automated software, which can be labor-intensive, difficult to reproduce, and sensitive to image quality and comet morphology. COMPASS automates comet assay image analysis by combining deep learning--based comet segmentation with damage measurement and automated comet selection. The pipeline produces standardized DNA damage measurements while offering robust detection, reducing manual effort and improving reproducibility through transparent selection and optional manual review. \\
\textbf{Availability and implementation:}
COMPASS is implemented in Python and is freely available at \url{https://github.com/rsinghlab/COMPASS}. Installation instructions, pretrained weights, and example usage are provided in the repository.\\
\textbf{Contact:}
jack\_roberts2@brown.edu,  
ritsingh@illinois.edu\\
\textbf{Supplementary information:}
Available at \textit{Bioinformatics} online.
}

\keywords{Comet Assay, Deep Learning, Segmentation}

\maketitle
\pagestyle{plain}
\thispagestyle{plain}

\section{Introduction}

DNA is the source of genetic information for all cellular life. It is constantly exposed to endogenous and exogenous factors that can cause lasting DNA damage, leading to potentially disadvantageous mutations, genomic instability, and diseases like cancer \citep{Chatterjee2017}. DNA damage analysis is therefore valuable for tasks such as understanding the genotoxicity of compounds, disease pathogenesis, and cellular aging. First developed by Singh et al.~\citeyearpar{Singh1988},
single-cell gel electrophoresis, known as the comet assay, requires only a small number of cells per sample and provides a highly sensitive way to measure DNA damage in a variety of forms \citep{Collins2004}.

Researchers can analyze the comet-like images from single-cell gel electrophoresis through visual inspection or commercially available software. When available, software is preferred for its reproducibility and increased throughput \citep{Kumaravel2009}. The earliest software solutions for comet scoring were manual, requiring user interaction to specify the location of each comet~\citep{Helma2000,Konca2003,Schwanz2011}. About a decade later, OpenComet~\citep{Gyori2014} offered an open-source software that leveraged intensity profile analysis and geometric heuristics for automatic comet detection and scoring. CometQ~\citep{Ganapathy2016} proposed an alternative automatic scoring pipeline by introducing a Support Vector Machine-based image classification step to adapt downstream thresholding and intensity-based segmentation to the type of image (silver or fluorescent stained) and level of DNA damage, at the cost of increased computational complexity. HiComet~\citep{Lee2018} offered rapid comet detection and scoring using histogram thresholding and clustering with a user-specified hyperparameter. 

The emergence of deep learning and convolutional neural networks (CNNs) demonstrated strong performance in computer vision tasks such as object detection and image segmentation \citep{Ren2015FasterRCNN}\citep{Redmon2016YOLO}\citep{He2017MaskRCNN}, making them a natural alternative to the heuristic comet analysis pipelines mentioned. The first application of deep learning to this task was DeepComet~\citep{Hong2020}, which used a Mask Region-based CNN architecture with an additional keypoint detection head for comet segmentation and scoring. The DeepComet authors also released the first publicly available labeled comet assay dataset of canine peripheral blood mononuclear cells (PBMCs). Later, GamaComet~\citep{Afiahayati2022} was released as a web-based tool for automatic comet scoring which uses the Faster R-CNN model trained on buccal mucosa cells. The most recent deep learning application we are aware of is a fully convolutional neural network ensemble for segmentation, followed by image post-processing and expert scoring of the refined image \citep{RuzSuarez2023}. 

These deep learning methods improve comet localization and segmentation accuracy relative to heuristic-based tools; however, in practice expert intervention is still required to \textit{select} specific comets for analysis. In particular, overlapping or partially truncated comets are typically excluded, as occluded or missing intensity information can lead to unreliable damage estimates. Moreover, comets exhibiting extreme or atypical damage may not be representative of the underlying cell population and are therefore often omitted to ensure robust and biologically meaningful measurements.

We present COMPASS, the \textbf{C}omet \textbf{O}bject \textbf{M}easurement \textbf{P}ipeline with \textbf{A}utomated \textbf{S}election and \textbf{S}coring. COMPASS replaces heuristic comet localization with a deep learning-based segmentation model and provides automated selection rules, auditable rejection flags, visual overlays, and optional manual review. COMPASS not only detects and scores comets, but also makes the user-facing decision of whether each comet should enter downstream analysis reproducible and reviewable. 

To evaluate these capabilities, we deploy COMPASS on the public DeepComet dataset (with ground-truth annotations) and on independently collected out-of-sample bat (\textit{Eptesicus fuscus}) and painted turtle (\textit{Chrysemys picta}) datasets collected as a part of a large cross-institutional effort to study aging across species. Details about the out-of-sample datasets are provided in Supplementary Materials section S4. We use `out-of-sample' to denote datasets completely held out from model training and development. On the public DeepComet benchmark, COMPASS achieves the strongest comet localization performance, improving bounding-box mean average precision (mAP$^{\mathrm{bb}}$@[0.5:0.95], averaged over intersection-over-union thresholds of 0.50–0.95) from 0.664 to 0.713 relative to DeepComet’s Mask R-CNN model. COMPASS also achieves competitive segmentation performance, improving mask mean average precision (mAP$^{\mathrm{mask}}$@[0.5:0.95]) from 0.584 to 0.593 on the challenging ``hard'' comet test split relative to DeepComet. Compared with OpenComet across 15 out-of-sample images, COMPASS reduces the total number of non-comet objects incorrectly detected as comets (false positives) from 130 to 6 while increasing correctly detected comets (true positives) from 128 to 146. Furthermore, its automated selection module, a capability not offered by OpenComet, achieves 0.986 precision and 0.858 recall.

Together, these capabilities make COMPASS an effective, open-source tool for comet assay analysis, reducing manual effort and supporting reproducible results across diverse experimental settings.

\section{Implementation}

\begin{figure}[t]
    \centering
    \includegraphics[width=\linewidth]{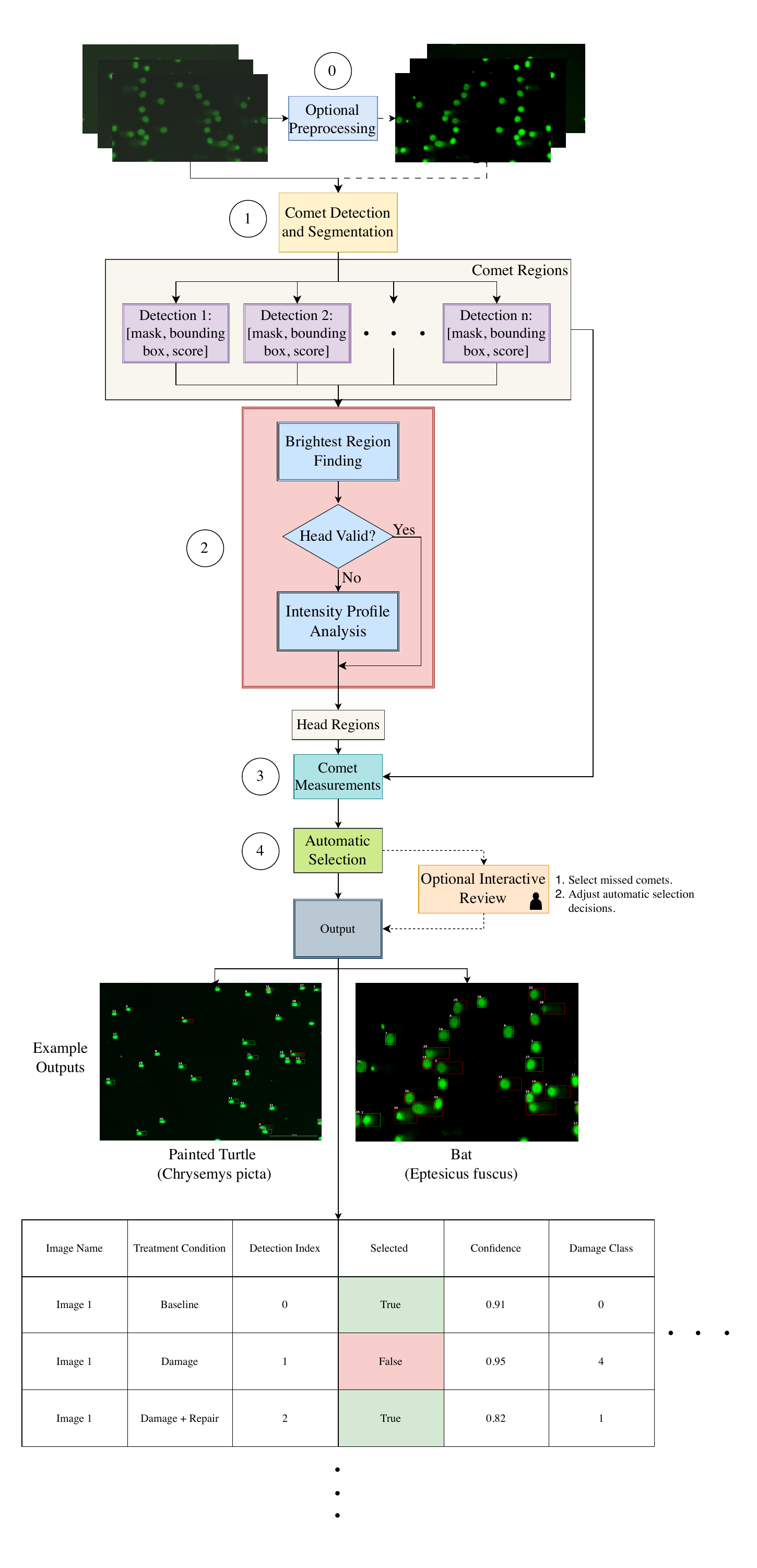}
    \caption{COMPASS pipeline visualisation.}
    \label{fig:1}
\end{figure}

COMPASS replaces OpenComet's threshold- and shape-based comet finding with a learned detection and segmentation model trained on the DeepComet dataset, while retaining the overall style of downstream head finding and measurement. The resulting pipeline can be seen in Figure ~\ref{fig:1} and has four main steps.

\subsection*{Step 0: Optional preprocessing for out-of-distribution images}

COMPASS accepts folders of TIFF (.tif/.tiff), PNG, JPEG (.jpg/.jpeg), and other Pillow-supported images, which it automatically converts to RGB for processing. However, for datasets whose raw image distribution differs substantially from the DeepComet-style images (PBMCs) used for model training, COMPASS also includes an optional preprocessing step. Users can assess the need for preprocessing by visually comparing their images with the training examples and inspecting detection outputs; poor detection on images with visibly different characteristics indicates that preprocessing may be beneficial. This optional preprocessing is detailed in Supplementary Materials section S3 and serves to improve robustness when image contrast, orientation, or color mapping differ significantly from the distribution of data the segmentation model was trained on.

\subsection*{Step 1: Detection and Segmentation}

Unlike OpenComet, which relies on thresholding and geometric heuristics for comet localization, COMPASS uses a deep learning--based instance segmentation model to detect and segment comets. By default, COMPASS employs a YOLOv11 segmentation model (or YOLO model)~\citep{yolo11_ultralytics} fine-tuned on the publically available DeepComet dataset, although COMPASS supports alternative finetuned segmentation backends which users can easily select with a command line argument. Section S2 of the Supplementary Materials provides more detail about the deep learning models and their training. 

The YOLO model predicts bounding boxes, instance masks, and confidence scores for each comet in a single pass. Before analysis, COMPASS filters detections to remove enclosed or duplicate instances before downstream analysis. Replacing OpenComet's handcrafted localization rules with a learned segmentation model substantially improves detection accuracy on both benchmark and independently collected datasets. These results are provided in Supplementary Material Tables S1,S2, and S3.

\subsection*{Step 2: Profile intensity and head finding}
For each retained detection, the model-predicted instance mask defines the comet region used for head finding. COMPASS performs head localization with the same approach as OpenComet, first attempting a brightest-region rule and falling back to profile-intensity analysis when the brightness-based head is not feasible. COMPASS also applies local column-wise background correction to eliminate noise before measurement.

\subsection*{Step 3: Scoring}
COMPASS reports the standard measurements, which can be seen along with their full descriptions in Supplementary Table S4.

\subsection*{Step 4: Selection and outputs}
COMPASS assigns each comet a rule-based damage class based on its tail DNA percentage, then filters comets using treatment-specific criteria and quality flags. Baseline images keep lower-damage classes (e.g., [0,1]), while damage/repair images keep intermediate classes (e.g., [1,2,3]). Users are able to adjust these rules as they see fit to account for differences in cell populations and lab processes. COMPASS also flags and excludes overlapping comets, comets cutoff at image borders, and low-signal or empty artifacts. As Figure~\ref{fig:1} shows, COMPASS produces a visualization of the selected comets with green boxes and the deselected comets with red boxes, together with a spreadsheet of the measurements, selection decision, user interactions, and factors informing each decision to ensure the entire automated process remains auditable.

\section{COMPASS Graphical User Interface}

COMPASS includes an optional OpenCV-based~\citep{opencv_library} graphical user interface (GUI) that opens after automatic detection, measurement, and selection. The interface presents one image at a time, allowing users to quickly verify automated decisions and make corrections as needed before COMPASS saves the final outputs. Clicking a model-predicted box toggles its selection state: users can manually reject automatically selected comets and manually select automatically rejected detections. Users can also drag to draw a new box around a missed comet. COMPASS records whether each final row originated from the model or from the user and stores the review action if one was taken. Final visuals preserve these distinctions: solid green marks automatically selected comets, solid red marks automatically rejected detections, dotted green marks manually selected detections, dotted red marks manually rejected detections, and blue marks manually added comets. This design allows users full control if they wish to edit the software's decisions or pick up any missed detections.

\section{Software Usage}

COMPASS is available at \url{https://github.com/rsinghlab/COMPASS}. The repository provides installation instructions, pretrained weights, and usage examples, while additional details regarding software usage, reproducibility, and output formats are available in Supplementary Material Section S1.

COMPASS uses a configurable detection confidence threshold to determine the minimum model confidence required for a predicted comet to be retained for downstream processing. Lower thresholds increase sensitivity by retaining more candidate detections but may introduce additional false positives, whereas higher thresholds provide more conservative detections. We recommend the default threshold of 0.2 for images similar to the DeepComet training data. For image sets that differ substantially from the training distribution, lowering the threshold may improve recall at the cost of precision. In our out-of-sample evaluation, we processed the painted turtle images using the default threshold of 0.2 and the bat images using an empirically determined threshold of 0.05 together with the optional preprocessing step.

COMPASS accepts standard image formats, including TIFF, PNG, and JPEG, and automatically resizes all input images to a fixed resolution of $960\times960$ pixels before model inference. Consequently, the deep learning inference stage is independent of the original image dimensions, although very large images may incur a small additional overhead ($<$ 0.1s) during file loading and resizing. Users therefore do not need to resize images prior to analysis.

\section{Validation}

We evaluated COMPASS through three complementary experiments. 

First, we evaluated detection and segmentation performance on the public DeepComet dataset, the only publicly available comet assay dataset accompanied by both expert annotations and open-source deep learning implementations. We evaluated the performance using mean Average Precision (mAP), the standard object detection and instance segmentation metric; mAP$^{\mathrm{bb}}$ denotes bounding-box mAP, while mAP$^{\mathrm{mask}}$ denotes mask-based segmentation mAP. COMPASS achieved the strongest overall localization performance, improving bounding-box detection accuracy from 0.664 to 0.713 mAP$^{\mathrm{bb}}$@[0.5:0.95] relative to DeepComet's Mask R-CNN model while maintaining competitive segmentation quality. On the challenging hard-comet subset, COMPASS also improved segmentation performance from 0.584 to 0.593 mAP$^{\mathrm{mask}}$@[0.5:0.95], demonstrating improved robustness on difficult examples. Detailed results and discussion are provided in Supplementary Material Section S5.

Second, we evaluated detection and automated selection performance on independently collected out-of-sample comet assay images from bats (\textit{Eptesicus fuscus}) and painted turtles (\textit{Chrysemys picta}). This evaluation used fifteen images in total (six bat images and nine painted turtle images) spanning baseline, UV damage, and UV damage plus repair conditions. Members of the laboratories that collected the data independently reviewed these images to establish reference labels, identifying both the comets that should be detected and those that should be retained for downstream analysis. We excluded these images from model development, including training and model selection. OpenComet outputs were also available, enabling direct comparison against the most widely used open-source comet analysis software. COMPASS substantially improved detection precision and reduced false positives relative to OpenComet (Table~\ref{tab:validation_summary}). Across all images, false positives (FP) across all fifteen images decreased from 130 with OpenComet to 6 with COMPASS, false negatives (FN) decreased from 27 to 14, and true positives (TP) increased from 128 to 146. We evaluated selection only on matched true-positive COMPASS detections, ensuring that the reported metrics reflect the quality of the retain/reject decision rather than localization errors. COMPASS achieved selection precision of 0.986 and recall of 0.858, indicating that the automated selection module reliably identifies biologically valid comets while minimizing inappropriate inclusions. Detailed per-image detection and selection results are provided in Supplementary Material Section S6.

Third, we evaluated measurement agreement between COMPASS and OpenComet. Because measurement validation benefits from larger numbers of matched comets, this analysis included five additional painted turtle images containing substantially denser comet populations (approximately 60 comets per image compared with approximately 10 in the selection dataset). We used these images exclusively for measurement validation and did not include them in the detection and selection evaluation. COMPASS demonstrated strong agreement with OpenComet on 141 matched comets across the five images while preserving the expected ordering of baseline, damage, and repair conditions. In particular, olive moment, one of the most widely used composite measures of DNA damage, showed a Pearson correlation of $r=$0.945 and a concordance correlation coefficient of 0.857. Although modest systematic differences in absolute values were observed, COMPASS consistently reproduced the treatment-dependent trends measured by OpenComet, suggesting that both tools would support the same biological conclusions across experimental groups. Detailed measurement agreement results are provided in Supplementary Material Section S7 (Table S5).

\begin{table}[h]
\centering
\caption{Out-of-sample validation summary. Detection counts are summed across all bat and painted turtle images; precision and recall are averaged per image. OpenComet does not automatically select comets hence the omitted results.}
\label{tab:validation_summary}
\begin{tabular}{lcc}
\toprule
Metric & OpenComet & COMPASS \\
\midrule
Detection TP/FP/FN & 128/130/27 & \textbf{146/6/14} \\
Detection precision & 0.636 & \textbf{0.964} \\
Detection recall & 0.845 & \textbf{0.899} \\
Selection TP/FP/FN & -- & 71/2/14 \\
Selection precision & -- & 0.986 \\
Selection recall & -- & 0.858 \\
\bottomrule
\end{tabular}
\end{table}

\section{Conclusion}

COMPASS provides an automated and auditable workflow for comet detection, measurement, selection, and scoring. By combining deep learning-based segmentation with standardized DNA damage measurement and automated comet selection, COMPASS provides accurate, automated, and transparent comet analysis. Validation on both publically available benchmark data and independently collected out-of-sample datasets demonstrates improved detection performance, reliable automated selection, and measurement agreement with established analysis software. Importantly, COMPASS does not remove the user from the loop entirely, allowing researchers to inspect, accept, reject, or add detections before COMPASS finalizes the results. This combination of automation, transparency, and user control makes COMPASS a practical tool for reproducible comet assay analysis across diverse experimental settings.

\par\noindent\textbf{Conflict of interest:} None declared.\par

\section{Funding}

This work is supported by the NSF Award 2213824 to the Biological Integration Institute: Integration Initiative: Sex, Aging, Genomics, \& Evolution (IISAGE). 

\section{Data availability}

The DeepComet dataset used for model development and benchmark evaluation is publicly available from the original publication \citep{Hong2020}. The bat and painted turtle images are available at \href{https://zenodo.org/records/20880511}{Zenodo}.

\section{Code availability}

COMPASS source code is freely available under the MIT license at \url{https://github.com/rsinghlab/COMPASS}.

\bibliographystyle{oup-abbrvnat}
\bibliography{ref}

@article{Chatterjee2017,
  author = {Chatterjee, Nimrat and Walker, Graham C.},
  title = {Mechanisms of DNA damage, repair, and mutagenesis},
  journal = {Environmental and Molecular Mutagenesis},
  year = {2017},
  volume = {58},
  number = {5},
  pages = {235--263},
  doi = {10.1002/em.22087}
}

@article{Collins2004,
  author = {Collins, Andrew R.},
  title = {The comet assay for DNA damage and repair},
  journal = {Molecular Biotechnology},
  year = {2004},
  volume = {26},
  number = {3},
  pages = {249--261},
  doi = {10.1385/MB:26:3:249}
}

@article{Kumaravel2009,
  author = {Kumaravel, T. S. and Vilhar, Barbara and Faux, Stephen P. and Jha, Awadhesh N.},
  title = {Comet Assay measurements: a perspective},
  journal = {Cell Biology and Toxicology},
  year = {2009},
  volume = {25},
  pages = {53--64},
  doi = {10.1007/s10565-007-9043-9}
}

@article{Schwanz2011,
  author  = {Schwanz, Lisa and Warner, Daniel A. and McGaugh, Suzanne and Di Terlizzi, Roberta and Bronikowski, Anne},
  title   = {State-dependent physiological maintenance in a long-lived ectotherm, the painted turtle ({C}hrysemys picta)},
  journal = {Journal of Experimental Biology},
  year    = {2011},
  volume  = {214},
  number  = {1},
  pages   = {88--97},
  doi     = {10.1242/jeb.046813},
  publisher = {The Company of Biologists Ltd}
}

@article{Singh1988,
  author  = {Singh, Narendra P. and McCoy, Michael T. and
             Tice, Raymond R. and Schneider, Edward L.},
  title   = {A Simple Technique for Quantitation of Low Levels of
             {DNA} Damage in Individual Cells},
  journal = {Experimental Cell Research},
  year    = {1988},
  volume  = {175},
  number  = {1},
  pages   = {184--191},
  month   = mar,
  doi     = {10.1016/0014-4827(88)90265-0},
  pmid    = {3345800},
  issn    = {0014-4827}
}

@article{Helma2000,
  author = {Helma, Christoph and Uhl, Michael},
  title = {A public domain image-analysis program for the single-cell gel electrophoresis (comet) assay},
  journal = {Mutation Research/Genetic Toxicology and Environmental Mutagenesis},
  year = {2000},
  volume = {466},
  number = {1},
  pages = {9--15},
  doi = {10.1016/S1383-5718(99)00232-6}
}

@article{Konca2003,
  author = {Ko{\'n}ca, Katarzyna and Lankoff, Anna and Banasik, Andrzej and Lisowska, Halina and Kuszewski, Tomasz and G{\'o}{\'z}d{\'z}, S{\l}awomir and Koza, Zbigniew and W{\'o}jcik, Andrzej},
  title = {A cross-platform public domain {PC} image-analysis program for the comet assay},
  journal = {Mutation Research/Genetic Toxicology and Environmental Mutagenesis},
  year = {2003},
  volume = {534},
  number = {1--2},
  pages = {15--20},
  doi = {10.1016/S1383-5718(02)00251-6}
}

@article{Gyori2014,
  author = {Gyori, Benjamin M. and Venkatachalam, Gokul and Thiagarajan, P. S. and Hsu, David and Clement, Marie-Veronique},
  title = {{OpenComet}: an automated tool for comet assay image analysis},
  journal = {Redox Biology},
  year = {2014},
  volume = {2},
  pages = {457--465},
  doi = {10.1016/j.redox.2013.12.020}
}

@article{Ganapathy2016,
  author = {Ganapathy, Subramanian and Muraleedharan, Anu and Savithri, S. P. and Chand, P. and Rajkumar, R. P.},
  title = {{CometQ}: an automated tool for the detection and quantification of {DNA} damage using comet assay image analysis},
  journal = {Computer Methods and Programs in Biomedicine},
  year = {2016},
  volume = {133},
  pages = {143--154},
  doi = {10.1016/j.cmpb.2016.05.020}
}

@article{Lee2018,
  author = {Lee, T. and Lee, S. and Sim, W. Y. and Jung, Y. M. and Han, S. and Won, J. H. and Min, H. and Yoon, S.},
  title = {{HiComet}: a high-throughput comet analysis tool for large-scale {DNA} damage assessment},
  journal = {BMC Bioinformatics},
  year = {2018},
  volume = {19},
  number = {1},
  pages = {49--61},
  doi = {10.1186/s12859-018-2015-7}
}

@inproceedings{Ren2015FasterRCNN,
  author = {Ren, Shaoqing and He, Kaiming and Girshick, Ross and Sun, Jian},
  title = {Faster {R-CNN}: Towards Real-Time Object Detection with Region Proposal Networks},
  booktitle = {Advances in Neural Information Processing Systems},
  year = {2015},
  volume = {28},
  pages = {91--99}
}

@inproceedings{Redmon2016YOLO,
  author = {Redmon, Joseph and Divvala, Santosh and Girshick, Ross and Farhadi, Ali},
  title = {You Only Look Once: Unified, Real-Time Object Detection},
  booktitle = {Proceedings of the IEEE Conference on Computer Vision and Pattern Recognition},
  year = {2016},
  pages = {779--788},
  doi = {10.1109/CVPR.2016.91}
}

@inproceedings{He2017MaskRCNN,
  author = {He, Kaiming and Gkioxari, Georgia and Doll{\'a}r, Piotr and Girshick, Ross},
  title = {Mask {R-CNN}},
  booktitle = {Proceedings of the IEEE International Conference on Computer Vision},
  year = {2017},
  pages = {2961--2969},
  doi = {10.1109/ICCV.2017.322}
}

@article{Hong2020,
  author = {Hong, Yeongeun and Han, Hye Jin and Lee, Hyun and others},
  title = {Deep learning method for comet segmentation and comet assay image analysis},
  journal = {Scientific Reports},
  year = {2020},
  volume = {10},
  pages = {18915},
  doi = {10.1038/s41598-020-75592-7}
}

@article{Afiahayati2022,
  author = {Afiahayati and Anarossi, Enenda and Yanuaryska, Reza Diandra and Mulyana, Sulaeman},
  title = {{GamaComet}: A deep learning-based tool for the detection and classification of {DNA} damage from buccal mucosa comet assay images},
  journal = {Diagnostics},
  year = {2022},
  volume = {12},
  number = {8},
  pages = {2002},
  doi = {10.3390/diagnostics12082002}
}

@article{RuzSuarez2023,
  author = {Ruz-Suarez, Daniel and Martin-Gonzalez, Ana and Brito-Loeza, Carlos and Uc-Cetina, Victor},
  title = {Fully-convolutional neural networks ensemble for comet segmentation in single cell gel electrophoresis assay images},
  journal = {Computer Methods in Biomechanics and Biomedical Engineering: Imaging \& Visualization},
  year = {2023},
  volume = {11},
  number = {5},
  pages = {1960--1972},
  doi = {10.1080/21681163.2023.2203279}
}

@misc{yolo11_ultralytics,
  author = {Jocher, Glenn and Qiu, Jing},
  title = {Ultralytics {YOLO11}},
  year = {2024},
  note = {Version 11.0.0},
  url = {https://github.com/ultralytics/ultralytics}
}

@article{opencv_library,
    author = {Bradski, G.},
    title = {The OpenCV Library},
    journal = {Dr. Dobb's Journal of Software Tools},
    year = {2000}
}

\end{document}